\documentclass{article} \usepackage{spconf,amsmath,graphicx}

\usepackage{hyperref}
\usepackage{multirow}

\title{Europarl-ST: A Multilingual corpus for Speech Translation of Parliamentary Debates}
\name{Javier Iranzo-S\'{a}nchez, Joan Albert Silvestre-Cerd\`{a}, Javier Jorge, Nahuel Rosell\'{o},}
\secondlinename{Adri\`{a} Gim\'{e}nez, Albert Sanchis, Jorge Civera, Alfons Juan \thanks{The research leading to these results has received funding from the European
		Union's Horizon 2020 research and innovation programme under grant agreement
		no. 761758 (X5gon); MCIU/AEI/FEDER,UE under the Multisub
		(RTI2018-094879-B-I00) research project and 
		the Government of Spain's FPU scholarship FPU18/04135.}}
\address{Machine Learning and Language Processing (MLLP) research group \\ Valencian Research Institute for Artificial Intelligence (VRAIN) \\ Universitat Polit\`{e}cnica de Val\`{e}ncia, Spain}

\begin{document}
%
\maketitle
\begin{abstract}

Current research into spoken language translation (SLT), or speech-to-text translation,
is often hampered by the
lack of specific data resources for this task, as currently available
SLT datasets are restricted to a limited set of language pairs. In this
paper we present \emph{Europarl-ST}, a novel multilingual SLT corpus
containing paired audio-text samples for SLT from and
into 6 European languages, for a total of 30 different translation
directions. This corpus has been compiled using the debates held 
in the European Parliament in the period between 2008 and 2012.
This paper describes the corpus creation process and presents a series
of automatic speech recognition, machine translation and spoken language translation 
experiments that highlight the potential of this new resource. The
corpus is released under a Creative Commons license and is freely
accessible and downloadable.

\end{abstract}

\section{Introduction} \label{sec:intro}

The significant developments in the automatic speech recognition (ASR)
and machine translation (MT) fields in the last five years, which have
been mainly driven by advances in deep learning models and greater
data availability, have picked up interest in spoken language translation (SLT)
as the natural convergence of the two previous fields.

However, SLT is far from solved. Two approaches are currently used:
cascade~\cite{sperber2017toward, DBLP:conf/interspeech/ChoNW17,
  Matusov18} and end-to-end models~\cite{DBLP:conf/interspeech/WeissCJWC17,sperber-etal-2019-attention,
  salesky-etal-2019-exploring}, without one being clearly adopted by
the community. The latest IWSLT 2018 evaluation campaign showed that
the cascade approach outperforms end-to-end models~\cite{iwslt2018},
but recent developments in the area are shrinking that gap~\cite{Jia2019}. The
performance of SLT, and especially end-to-end SLT models, is limited by
the lack of SLT corpora when compared with the more resource-rich ASR
and MT fields.
Furthermore, most of the existing SLT corpora are limited to only
English speech data paired with translations into other languages,
such as the recently released MuST-C
corpus~\cite{DBLP:conf/naacl/GangiCBNT19}.
This fact limits the SLT research than could be carried out in language pairs other than English.  Moreover, recent studies report their
main results using either the paid Fisher/Callhome corpora~\cite{sperber2017toward,
  DBLP:conf/interspeech/WeissCJWC17,sperber-etal-2019-attention,
  salesky-etal-2019-exploring,post2013improved}, or private propietary
datasets~\cite{Jia2019}, which limits reproducibility for the research
community.


In order to alleviate these problems, we have created the
\emph{Europarl-ST} corpus out of European Parliament (EP) debates and their
official transcriptions and translations.  To our knowledge,
Europarl-ST is the first fully self-contained, 
publicly available corpus with both, multiple (speech) source
and target languages, which will also enable further research into
multilingual SLT (cf.~\cite{boito2020mass}). The Europarl-ST corpus is released under a Creative
Commons Attribution-NonCommercial 4.0 International license (CC BY-NC
4.0), and can be freely accessed and downloaded
at \\
\url{www.mllp.upv.es/europarl-st}.

\section{Data collection and processing} \label{sec:data}

The corpus has been created using the publicly available videos from European
Parliament
debates\footnote{http://www.europarl.europa.eu/plenary/en/debates-video.html}. %
In order to ease the access to the different attributes of each debate 
the LinkedEP database is used~\cite{van2017debates}. 
The basic unit of this corpus is a \textit{speech}, an intervention
made by a single speaker at the Parliament. 

The EP debates suffer from missing videos, inaccurate timestamps and,
as of 2011, many translations into languages other than English are
missing. Indeed, after 2012, the translation of EP debates is not
available. Additional data is discarded when constructing the
Europarl-ST corpus, since in order to build a corpus of
audio-transcription-translation triples, it is necessary to properly
define forced audio-text and text-text sentence alignments, and
intra-sentence word-alignment.

For this initial release of the corpus, experiments are reported from
and into English (En), German (De), French (Fr) and Spanish (Es), since these languages
accumulate a larger number of speech hours.
Additional languages, such as Italian and Portuguese, 
will also be included in the initial release, but experimental 
results are not reported due to time constraints.

\subsection{Audio-to-text alignment and data filtering} 


One of the challenges processing this corpus is that timestamps
provided for the EP speeches can be wildly inaccurate, and as a
side-effect, they often contain fragments from both the preceding and
following speeches.
In order to ameliorate this, first we carried out a Speaker
Diarization (SD) step for each speech using the \textit{LIUM
  SpkDiarization} \cite{DBLP:conf/interspeech/RouvierDGKMM13}
toolkit. Second, for each speech, the longest sequence of audio
segments belonging to the same speaker was clipped,
making the assumption that it does correspond to the actual intervention of the
speaker of this speech. 
Finally, a forced alignment of the clipped audio segments was carried
out against their corresponding transcriptions to obtain correct word
timestamps.  Forced alignments were carried out using the TLK toolkit's
decoder~\cite{Agua14} and the FF-DNN acoustic models (AM) described in
Section~\ref{sec:exp-setup-asr}, restricting the search graph of the
decoder to the provided transcription. As a result of the procedure
describe above (Step 1), around 28\% of the original audio data was
discarded (see Table~\ref{tab:raw_stats} for language-based
statistics).

\begin{table}
\caption{Number of speech hours after each step of the data filtering pipeline,
and CER of the filtered data sets.} \label{tab:raw_stats} 
\centering

\begin{tabular}{l|rrr|r}
 & Initial & Step 1 & Step 2 & CER   \\ \hline
De & 207 & 149   & 44     & 10.7 \\
En & 346 & 252   & 120    & 12.9  \\
Es & 80  & 59    & 34     & 9.1   \\
Fr & 183 & 132   & 47     & 10.7  \\

\end{tabular} 
\addtolength{\tabcolsep}{4pt}    
\end{table}


%


Next, in order to produce a reliable corpus than could be used to both
train and evaluate models, a second data filtering step was carried
out based on character error rate (CER) computed at the speech
level. First, we apply ASR over all speeches, using the ASR system
described in Section~\ref{sec:exp-setup-asr}.  Second, we measure how
much the recognition outputs differ from the provided reference
transcriptions by computing CER values. Our aim is to eliminate
speeches that exhibit significant amounts of non-verbatim
transcriptions, as well as non-transcribed speech or unuttered
transcripts that could be present either due to mistakes of the SD
process or to annotation errors in the original data.  In comparison
with the well-known word error rate (WER) metric, the CER is more
convenient for our purposes, as it better gauges the phonetic
similarity between the recognised speech and the candidate reference
transcripts, and alleviates the effect of ASR out-of-vocabulary words.

Finally, language-dependent CER thresholds were defined, 15\% for
French, German and Spanish and 20\% for English, in order to exclude
those speeches whose CER exceeded these thresholds.  Thresholds were
defined based on previous experience filtering crawled speech data.
As a result of this filtering step (Step 2), around 40-70\% of the
audio data selected in the previous step was discarded (see
Table~\ref{tab:raw_stats} for detailed statistics).  CER figures
computed on the selected speeches after Step 2 are also provided in
Table~\ref{tab:raw_stats}.  These figures are an approximation to a
quality assurance measure to ensure that only speeches with little or
no noise are included into the corpus.  At the end of this process,
around 60-80\% of the original data was filtered out.


\subsection{Source-to-target text alignment} 

Each selected speech, both transcription and translation, is divided into 
sentences, using the \textit{sentence-split.pl} script from the Moses
toolkit~\cite{DBLP:conf/acl/KoehnHBCFBCSMZDBCH07}, that are aligned
using Gargantua~\cite{DBLP:conf/coling/BrauneF10}. 
Sentences longer than 20 seconds were split into shorter ones in order
to accommodate the data for training purposes. Shorter sentences were
generated by computing word-alignments using
Fast-align~\cite{DBLP:conf/naacl/DyerCS13} and pairing them to
guarantee intra-sentence alignments.  The statistics of the remaining
data after text-aligning and excluding speeches with no translation
into the respective target language are shown in
Table~\ref{tab:corpus_stats}.  As observed in
Table~\ref{tab:corpus_stats}, this corpus is provided with
segmentations, both at the speech and sentence level.  The
sentence-level segmentation is expected to be devoted to training
purposes, while evaluations at the speech level are reported in
Section~\ref{sec:exp}.
 
\begin{table} \caption{Statistics of the preprocessed Europarl-ST corpus.} \label{tab:corpus_stats} 
\centering
\begin{tabular}{l@{~~}lrrrrr} 
Src & Trg & Speeches & Sent. & Hours & Src w. & Trg w. \\  \hline
\multirow{3}{*}{De} & En & 1521 & 18.1K & 42 & 345K & 409K  \\
& Es & 863 & 10.2K & 24 & 196K & 242K \\
& Fr & 839 & 9.6K & 24 & 191K & 265K \\ \hline
\multirow{3}{*}{En} & De & 3233 & 35.5K & 89 & 811K & 793K \\
 & Es & 3184 & 34.4K & 87 & 796K & 865K \\
 & Fr & 3174 & 34.5K & 87 & 794K & 974K \\ \hline
\multirow{3}{*}{Es} & De & 694 & 7.0K & 20 & 193K & 186K \\
& En & 1131 & 11.2K & 32 & 305K & 307K \\
& Fr & 684 & 6.9K & 20 & 190K & 225K \\ \hline
\multirow{3}{*}{Fr} & De & 832 & 9.6K & 25 & 263K & 227K \\
& En & 1306 & 15.1K & 38 & 394K & 371K \\
& Es & 817 & 9.4K & 25 & 260K & 246K \\
\end{tabular} \end{table}

A speaker-independent train/dev/test partition was defined, devoting
approximately 3 hours of audio to each of the dev and test sets, and
the rest was left as training data. The dev/test speakers are the same
for language directions with the same source language. However, the
number of speeches may differ because for some speeches there are
translations missing.  The training data might be used to fine-tune
and adapt out-of-domain models to this specific domain, or even to
train basic in-domain ASR, MT and SLT models from scratch.

\section{Experiments and results} \label{sec:exp}


This section introduces the setup used for the experiments performed
with the Europarl-ST corpus. In addition to ASR and MT experiments, SLT
experiments following a cascade approach, in which the output of an
ASR system is used as input for an MT system, are reported. First, the
performance of models trained on general domain data when applied to
the Europarl-ST corpus are evaluated, and second, the
usefulness of the Europarl-ST training data for adapting models
to the EP specific domain is also assessed.
More precisely, results of ASR, MT and SLT experiments are reported
using the 4 selected languages (English, German, Spanish and French),
for a total of 12 translation directions in the case of translation
experiments. Results are reported in terms of WER for ASR experiments,
and BLEU~\cite{DBLP:conf/acl/PapineniRWZ02} for MT and SLT experiments.

In order to properly compute BLEU, both the system hypothesis and the
reference translation must have the same number of lines. However, in
a SLT experiment, the number of lines will depend on the segmentation
applied to the output of the ASR system in the cascade case, and the
SLT system in the end-to-end case. Therefore, it is standard to
re-segment the system hypothesis in order to get the same number of
lines as in the reference.  This re-segmentation is performed with the
\textit{mwerSegmenter}~\cite{DBLP:conf/iwslt/MatusovLBN05}, and then
evaluated by computing case-sensitive BLEU (including punctuation
signs) with SacreBLEU~\cite{W18-6319}. All evaluations are carried out
at the speech level, so re-segmentation is applied to both, MT
and SLT experiments, in order to evaluate them under the same
conditions.

\subsection{ASR} \label{sec:exp-setup-asr}

General-purpose ASR systems for German (De), English (En), Spanish
(Es) and French (Fr) were used to generate automatic transcripts for
audio speeches in the development and test sets of each language
pair. These automatic transcripts are the input text for subsequent MT
systems within the SLT cascade approach.

These ASR systems are based on the hybrid deep neural network hidden
Markov model (DNN-HMM) approach. Acoustic models, are generated using
the TLK toolkit~\cite{Agua14} to train feed-forward (FF) DNN-HMM
models of three left-to-right tied triphone states, using 48 (De, Es,
Fr) or 80-dimensional (En) Mel frequency cepstral coefficients (MFCCs)
as input features. State tying was done by applying language-dependent
classification and regression trees (CART), which resulted in 10K (Es,
Fr) or 18K (De, En) tied triphone states.  With the exception of the
French ASR system which only features FF-DNNs, these models were used
to bootstrap bidirectional long-short term memory (BLSTM) DNN models,
the latter model trained using Tensorflow~\cite{tensorflow}. For
German, Spanish and French, we also trained fCMLLR AMs, so that these
systems follow a two-step recognition process.

On the other hand, regarding the language models (LM), we used a linear
combination of several $n$-gram LMs trained with SRILM~\cite{stolcke02},
combined with a recurrent NN (RNN) LM trained using the RNNLM
toolkit~\cite{rnnlm} (De, Es, Fr), or an LSTM LM trained with the CUED-RNNLM
toolkit~\cite{Chen2016} (En).  The vocabulary of these systems was restricted
to 200K words.  Table~\ref{tab:asr-resources-stats} shows overall statistics of
the amount of training data that were used to train the acoustic models, in
terms of speech hours, and the language models, in terms of sentences and
words. The number of English words includes 294G words from Google Books counts.

\begin{table} \caption{Statistics of AM and LM training data.} 
\label{tab:asr-resources-stats} 
\centering 
\begin{tabular}{l|c|cc}
 & Hours (K)   & Sentences (M) & Words (G)   \\\hline
De & 0.9  & 71 & 0.8    \\ 
En & 5.6  & 532 & 300  \\ 
Es & 0.8  & 24  & 0.7   \\ 
Fr & 0.7  & 110 & 1.8  \\
\end{tabular} \end{table}
 
Table~\ref{tab:asr} shows, for each SLT test set, WER figures computed from the ASR
part only. Rows represent source (ASR)
languages, whilst columns represent target (MT) languages. It is
important to remind that the set of source speeches, though mostly
overlapping, are different because the correspoding target text translation may
not exist.
Results show that most WER figures are below 20\%, except in those
pairs having French as input language. This is explained because the
French ASR system does not feature BLSTM acoustic models, and it is
the language with least acoustic resources.


\begin{table} \caption{ASR results in terms of WER on the test sets.} 
\label{tab:asr} 
\centering 
\begin{tabular}{l|cccc}
 & De & En & Es & Fr  \\\hline
De & -- & 19.8  & 19.8 & 19.9  \\ 
En & 17.2 & -- & 17.2 & 17.1 \\ 
Es & 14.6 & 15.0 & -- & 14.6 \\ 
Fr & 27.3 & 24.3 & 27.2 & -- \\
 \end{tabular} \end{table}
 
 \vspace{-2pt}

\subsection{MT} \label{sec:exp-setup-mt}

A Neural Machine Translation (NMT) system was built for each translation direction 
mainly using publicly available corpora from OPUS~\cite{DBLP:conf/lrec/Tiedemann12} and excluding the Europarl corpus to
avoid data overlapping. The training data used in each language pair is
shown in Table~\ref{tab:mt_training_data}. This includes the list 
of corpora and the total number of sentences.

\begin{table} \caption{Training data used for the MT systems} \label{tab:mt_training_data} 
\centering
\begin{tabular}{ccc} 
Pair & Corpora & \# sents(M) \\ \hline
\multirow{2}{*}{De$\leftrightarrow$En}  & DGT,eubookshop & \multirow{2}{*}{21.0} \\
& TildeMODEl, Wikipedia &  \\ \hline
\multirow{2}{*}{De$\leftrightarrow$Es} & DGT, eubookshop,  & \multirow{2}{*}{14.3} \\
& JRC-Acquis, TildeModel & \\  \hline
\multirow{2}{*}{De$\leftrightarrow$Fr} & eubookshop, JRC-Acquis,  & \multirow{2}{*}{14.3} \\
& TildeModel & \\ \hline
\multirow{2}{*}{En$\leftrightarrow$Es} & commoncrawl, eubookshop,  & \multirow{2}{*}{21.1} \\
& EU-TT2, UN, Wikipedia & \\ \hline
\multirow{2}{*}{En$\leftrightarrow$Fr} & commoncrawl, giga,  &  \multirow{2}{*}{38.2} \\
& undoc, news-commentary & \\ \hline
\multirow{2}{*}{Es$\leftrightarrow$Fr}  & DGT, eubookshop,  & \multirow{2}{*}{37.2} \\
& JRC-Acquis, UNPC & \\
\end{tabular} 
\end{table}


The corpora were preprocessed by applying 40K 
BPE~\cite{DBLP:conf/acl/SennrichHB16a} operations, learnt jointly over
the source and target data.  The models follow the Transformer NMT
architecture~\cite{DBLP:conf/nips/VaswaniSPUJGKP17} and are trained
using the Transformer BASE configuration using 4GPU machines and an
initial learning rate of $5\mathrm{e}{-4}$, decayed using the inverse
square root scheme. Once the training converges, a
fine-tuning step was carried out using the training data generated 
in Section~\ref{sec:data}. To do so, we fix the learning rate to
$5\mathrm{e}{-5}$, and we use a standard SGD optimizer instead of
Adam. We measure performance on the dev set and stop training once the
perplexity stops decreasing. 
Table~\ref{tab:mt_referenceTranscription} shows BLEU scores of the 
out-of-domain MT systems compared with those
obtained by fine-tuning with the Europarl-ST training data 
shown between parenthesis. These MT
systems are evaluated on automatic outputs generated from reference 
transcriptions as a standalone MT task.

\begin{table} \caption{BLEU scores of out-of-domain MT systems with
    reference transcriptions as input and fine-tuning BLEU scores between parenthesis.}
\label{tab:mt_referenceTranscription} \centering 
\addtolength{\tabcolsep}{-4pt}
\begin{tabular}{l|cccc}


  & De & En & Es & Fr  \\\hline
De & -- & 32.6 (\textbf{36.3})  & 26.8 (\textbf{29.3}) & 23.2 (\textbf{27.1}) \\ 
En & 33.6 (\textbf{37.6}) & -- & 46.3 (\textbf{48.2}) & 34.7 (\textbf{39.2}) \\ 
Es & 20.9 (\textbf{24.8}) & 39.2 (\textbf{41.8}) & -- & 29.3 (\textbf{33.1}) \\ 
Fr & 23.3 (\textbf{26.3}) & 38.7 (\textbf{42.3}) & 34.8 (\textbf{36.3}) & -- \\
 \end{tabular} 
\addtolength{\tabcolsep}{4pt}
\end{table}

The results vary depending on the amount of resources used for each
system as well as the intrinsic difficulty of each translation direction. 
As observed, the fine-tuned systems trained on the Europarl-ST
corpus provide
very significant improvements over the out-of-domain systems, ranging from
+1.9 up to +4.0 BLEU, which confirms the quality and usefulness of the
training data.

\vspace{-6pt} 
\subsection{SLT}

This section presents the results of the SLT experiments following the cascade
approach, in which we use the output of the ASR system as input for the MT
system. The output of the ASR system is segmented based on detected
silences. For this task, we will combine the ASR and MT models
described in Sections~\ref{sec:exp-setup-asr} and~\ref{sec:exp-setup-mt}.  
We use the fine-tuned MT systems as they
outperform the out-of-domain systems in all cases. 
The results of the SLT experiments are shown in Table~\ref{tab:slt}.

\begin{table} \caption{BLEU scores of cascade-based SLT experiments 
with fine-tuned models assessed on the test sets.} 
\label{tab:slt} \centering \begin{tabular}{l|cccc} 
  & De & En & Es & Fr  \\\hline
De & -- & 21.3 & 17.5 & 15.7 \\ 
En & 22.4 & --  & 28.0 & 23.4 \\ 
Es & 15.6  & 26.5 & -- & 22.0 \\ 
Fr & 15.3 & 25.4 & 23.2 & -- \\ 
\end{tabular} \end{table}

%
%
%

Table~\ref{tab:slt} shows that BLEU scores 
in the SLT experiments are lower than 
those in the MT experiments. This is to be expected,
as the MT system has to cope not only with error propagation from incorrect
transcriptions, but also with a sub-optimal segmentation of the input which might not
correspond with whole sentences. This could be improved with a specific segmentation
and punctuation module~\cite{DBLP:conf/interspeech/ChoNW17}.
As expected, although the overall BLEU scores are lower, the ranking of the performance across translation
directions is preserved, with MT systems that obtained the highest scores in the MT
experiments, also obtaining the highest scores in the SLT experiments, and vice versa. 
Although SLT results are constrained by the complexity of this task, these results serve
as a good starting baseline for future developments.


\section{Conclusions} \label{sec:conclus}

We have presented a novel SLT corpus built from European
Parliament proceedings. The experiments presented have shown how our
proposed filtering pipeline is able to extract good quality data that is useful
both for evaluating the performance of out-of-domain systems in this
task, as well as for system adaptation to the specific domain of
parliamentary debates.  We believe that the release of this
multi-source and multi-target corpus will enable further research into
multilingual SLT.

In terms of future work, the presented filtering pipeline can be
extended to cover additional languages in the future. Additionally, we
will study new filtering techniques to increase the amount of
hours available per each language pair.

Finally, we also plan on gauging the performance of end-to-end models for
this task, and compare it with cascade systems that use MT models adapted to the 
translation of ASR output. This adaptation can be carried out by training MT systems 
on real ASR output as source input~\cite{DBLP:conf/iwslt/PeitzWNN12} or on
simulated ASR output by applying noising techniques to the source side~\cite{sperber2017toward}.




\bibliographystyle{IEEEbib} \bibliography{bibliography}

\begin{thebibliography}{10}

\bibitem{sperber2017toward}
Matthias Sperber, Jan Niehues, and Alex Waibel,
\newblock ``Toward robust neural machine translation for noisy input
  sequences,''
\newblock in {\em IWSLT 2017}.

\bibitem{DBLP:conf/interspeech/ChoNW17}
Eunah Cho, Jan Niehues, and Alex Waibel,
\newblock ``{NMT-Based Segmentation and Punctuation Insertion for Real-Time
  Spoken Language Translation},''
\newblock in {\em Interspeech 2017}.

\bibitem{Matusov18}
E.~Matusov, P.~Wilken, P.~Bahar, J.~Schamper, P.~Golik, A.~Zeyer, J.A.
  Silvestre-Cerd{\`a}, A.~Mart{\'\i}nez-Villaronga, H.~Pesch, and J.~Peter,
\newblock ``{Neural Speech Translation at AppTek},''
\newblock in {\em IWSLT 2018}.

\bibitem{DBLP:conf/interspeech/WeissCJWC17}
Ron~J. Weiss, Jan Chorowski, Navdeep Jaitly, Yonghui Wu, and Zhifeng Chen,
\newblock ``Sequence-to-sequence models can directly translate foreign
  speech,''
\newblock in {\em Interspeech 2017}.

\bibitem{sperber-etal-2019-attention}
Matthias Sperber, Graham Neubig, Jan Niehues, and Alex Waibel,
\newblock ``Attention-passing models for robust and data-efficient end-to-end
  speech translation,''
\newblock {\em Transactions of the Association for Computational Linguistics},
  vol. 7, pp. 313--325, Mar. 2019.

\bibitem{salesky-etal-2019-exploring}
Elizabeth Salesky, Matthias Sperber, and Alan~W Black,
\newblock ``Exploring phoneme-level speech representations for end-to-end
  speech translation,''
\newblock in {\em {ACL} 2019}.

\bibitem{iwslt2018}
Jan Niehues, Roldano Cattoni, Sebastia Stüker, Mauro Cettolo, Marco Turchi,
  and Marcello Federico,
\newblock ``{The IWSLT 2018 Evaluation Campaign},''
\newblock in {\em {IWSLT} 2018}.

\bibitem{Jia2019}
Ye~Jia, Ron~J. Weiss, Fadi Biadsy, Wolfgang Macherey, Melvin Johnson, Zhifeng
  Chen, and Yonghui Wu,
\newblock ``{Direct Speech-to-Speech Translation with a Sequence-to-Sequence
  Model},''
\newblock in {\em Interspeech 2019}.

\bibitem{DBLP:conf/naacl/GangiCBNT19}
Mattia Antonino~Di Gangi, Roldano Cattoni, Luisa Bentivogli, Matteo Negri, and
  Marco Turchi,
\newblock ``{MuST-C: a Multilingual Speech Translation Corpus},''
\newblock in {\em {NAACL-HLT} 2019}.

\bibitem{post2013improved}
Matt Post, Gaurav Kumar, Adam Lopez, Damianos Karakos, Chris Callison-Burch,
  and Sanjeev Khudanpur,
\newblock ``{Improved speech-to-text translation with the Fisher and Callhome
  Spanish--English speech translation corpus},''
\newblock in {\em {IWSLT} 2013}.

\bibitem{boito2020mass}
Marcely~Zanon Boito, William~N. Havard, Mahault Garnerin, ´{E}ric Le~Ferrand,
  and Laurent Besacier,
\newblock ``Mass: A large and clean multilingual corpus of sentence-aligned
  spoken utterances extracted from the bible,''
\newblock in {\em LREC 2020 (accepted)}.

\bibitem{van2017debates}
Astrid Van~Aggelen, Laura Hollink, Max Kemman, Martijn Kleppe, and Henri
  Beunders,
\newblock ``{The debates of the European Parliament as linked open data},''
\newblock {\em Semantic Web}, vol. 8, no. 2, pp. 271--281, 2017.

\bibitem{DBLP:conf/interspeech/RouvierDGKMM13}
Mickael Rouvier, Gr{\'{e}}gor Dupuy, Paul Gay, Elie el~Khoury, T{\'{e}}va
  Merlin, and Sylvain Meignier,
\newblock ``An open-source state-of-the-art toolbox for broadcast news
  diarization,''
\newblock in {\em Interspeech 2013}.

\bibitem{Agua14}
Miguel~A. del Agua, Adri{\`{a}} Gim{\'{e}}nez, Nicol{\'{a}}s Serrano,
  Jes{\'{u}}s Andr{\'{e}}s{-}Ferrer, Jorge Civera, Alberto Sanch{\'{\i}}s, and
  Alfons Juan,
\newblock ``{The Translectures-UPV Toolkit},''
\newblock in {\em IberSpeech 2014}.

\bibitem{DBLP:conf/acl/KoehnHBCFBCSMZDBCH07}
Philipp~Koehn et~al.,
\newblock ``Moses: Open source toolkit for statistical machine translation,''
\newblock in {\em {ACL} 2007}.

\bibitem{DBLP:conf/coling/BrauneF10}
Fabienne Braune and Alexander~M. Fraser,
\newblock ``Improved unsupervised sentence alignment for symmetrical and
  asymmetrical parallel corpora,''
\newblock in {\em {COLING} 2010}.

\bibitem{DBLP:conf/naacl/DyerCS13}
Chris Dyer, Victor Chahuneau, and Noah~A. Smith,
\newblock ``A simple, fast, and effective reparameterization of {IBM} model
  2,''
\newblock in {\em {NAACL-HLT} 2013}.

\bibitem{DBLP:conf/acl/PapineniRWZ02}
Kishore Papineni, Salim Roukos, Todd Ward, and Wei{-}Jing Zhu,
\newblock ``{Bleu: a Method for Automatic Evaluation of Machine Translation},''
\newblock in {\em {ACL} 2002}.

\bibitem{DBLP:conf/iwslt/MatusovLBN05}
Evgeny Matusov, Gregor Leusch, Oliver Bender, and Hermann Ney,
\newblock ``Evaluating machine translation output with automatic sentence
  segmentation,''
\newblock in {\em {IWSLT} 2005}.

\bibitem{W18-6319}
Matt Post,
\newblock ``A call for clarity in reporting {BLEU} scores,''
\newblock in {\em WMT18}.

\bibitem{tensorflow}
``Tensorflow,'' \url{https://www.tensorflow.org/}.

\bibitem{stolcke02}
A.~Stolcke,
\newblock ``{SRILM -- an extensible language modeling toolkit},''
\newblock Denver, CO, USA, Sept. 2002, pp. 901--904.

\bibitem{rnnlm}
``{The RNNLM Toolkit},'' \url{http://www.fit.vutbr.cz/~imikolov/rnnlm/}.

\bibitem{Chen2016}
Xi~Chen, Xin Liu, Y.~Qian, Mark J.~F. Gales, and Philip~C. Woodland,
\newblock ``{CUED-RNNLM — An open-source toolkit for efficient training and
  evaluation of recurrent neural network language models},''
\newblock in {\em ICASSP 2016}.

\bibitem{DBLP:conf/lrec/Tiedemann12}
J{\"{o}}rg Tiedemann,
\newblock ``Parallel data, tools and interfaces in {OPUS},''
\newblock in {\em {LREC} 2012}.

\bibitem{DBLP:conf/acl/SennrichHB16a}
Rico Sennrich, Barry Haddow, and Alexandra Birch,
\newblock ``Neural machine translation of rare words with subword units,''
\newblock in {\em {ACL} 2016}.

\bibitem{DBLP:conf/nips/VaswaniSPUJGKP17}
Ashish Vaswani, Noam Shazeer, Niki Parmar, Jakob Uszkoreit, Llion Jones,
  Aidan~N. Gomez, Lukasz Kaiser, and Illia Polosukhin,
\newblock ``Attention is all you need,''
\newblock in {\em NIPS 2017}.

\bibitem{DBLP:conf/iwslt/PeitzWNN12}
Stephan Peitz, Simon Wiesler, Markus Nu{\ss}baum{-}Thom, and Hermann Ney,
\newblock ``Spoken language translation using automatically transcribed text in
  training,''
\newblock in {\em {IWSLT} 2012}.

\end{thebibliography}

\end{document}